\documentclass[conference]{IEEEtran}
\usepackage{cite}
\usepackage{amsmath,amssymb,amsfonts}
\usepackage{graphicx}
\usepackage{textcomp}
\usepackage{xcolor}
\usepackage{subcaption}
\usepackage{array}
\usepackage{hyperref}
\usepackage{booktabs}

\begin{document}

\title{Complexity Induction: Compositional Generalization via Structured Training Distortion}

\author{\IEEEauthorblockN{Aleksandr V. Abramov}
\IEEEauthorblockA{Independent Researcher \\
avabr.me@gmail.com}
}

\maketitle

\begin{abstract}
We demonstrate that structured distortion of training data --- which we term \textit{complexity induction} --- can induce compositional generalization in a standard CNN classifier without architectural modification. Using synthetic images of colored geometric shapes, we encode classes as flat string labels (e.g., ``red-circle'') with no explicit attribute decomposition, and exclude certain color-shape combinations from training entirely. We apply two distortion methods derived from Jaccard string similarity between class names: \textit{mixed labels} (soft target distributions encoding inter-class overlap) and \textit{expanded dataset} (false training samples with structurally motivated incorrect labels). Both methods induce the ability to predict unseen class combinations, and act at different levels: mixed labels activate the classifier for unseen combinations by exploiting the CNN's natural embedding structure, while expanded training improves the embedding factorization itself. A control with random (unstructured) false labels confirms that the effect depends on the structure of the distortion, not on noise per se. These results suggest that structured complication of training signals can influence both the internal organization of learned representations and their compositional interpretation --- a principle that may underlie the role of natural language in cognitive development.
\end{abstract}


\section{Introduction}

Modern machine learning operates on a fundamental assumption: training data should represent the target task as accurately as possible. Clean labels, balanced classes, and faithful correspondence between inputs and targets are considered prerequisites for successful learning. Techniques such as data augmentation, label smoothing, and curriculum learning refine this paradigm but do not challenge its core premise --- that the training signal should guide the model toward the correct mapping.

This paper investigates a different premise. We ask whether \textit{deliberately distorting} the training signal --- through structured modification of labels and injection of false training samples --- can induce qualitatively new capabilities in a standard classifier. Specifically, we examine whether a model trained on such distorted data can predict class combinations it has never seen, a property known as compositional generalization.

The motivation for this investigation comes from an observation about natural language. In early human development, language accompanies perception as an additional signal that is combinatorial, arbitrary, and partially misleading. A child hearing ``the cat came'' and ``the storm came'' encounters phonetically similar phrases referring to entirely unrelated events. Rather than simplifying perception, language \textit{complicates} it --- creating spurious associations between unrelated objects and contexts. Yet this complication appears to be productive: it is precisely in the presence of language that human cognition develops the capacity for compositional, generative thought.

We refer to this principle as \textit{complexity induction} --- the deliberate injection of external structural information into a parametric approximator through intentional distortion of training data. The core hypothesis is that such distortion, when structured rather than random, can induce compositional generalization absent from undistorted training.

To test this hypothesis in its simplest form, we design an experiment with the following properties:
\begin{itemize}
    \item \textbf{Synthetic data}: images of geometric shapes characterized by color and form, giving full control over object properties.
    \item \textbf{Flat class space}: classes are encoded as monolithic string labels (e.g., ``red-circle''), not as factored attribute pairs. The model receives no architectural or data-level hint that classes have internal structure.
    \item \textbf{Standard architecture}: a CNN feature extractor with a single linear classifier, enabling direct analysis of embedding factorization.
    \item \textbf{Holdout combinations}: certain color-shape pairs are excluded from training entirely and used only for evaluation.
\end{itemize}

We propose two complementary distortion methods derived from string similarity between class names. \textit{Mixed labels} replace one-hot targets with soft distributions encoding inter-class overlap (e.g., ``red-circle'' and ``red-triangle'' share ``red''). \textit{Expanded dataset} adds false copies of training images with structurally motivated incorrect labels. Both methods model, in simplified form, the way natural language creates associations between categories through shared symbolic components.

\section{Related Work}

Our work intersects several established research directions. We position it relative to each, highlighting the key distinction: existing methods either decompose class structure explicitly or use soft labels to improve within-distribution performance, whereas our approach uses structured label distortion to induce compositional generalization on a flat, undecomposed class space.

\subsection{Compositional Zero-Shot Learning}

Compositional Zero-Shot Learning (CZSL) addresses the task of recognizing unseen attribute-object combinations by learning from seen compositions \cite{czsl_survey_2025}. Current methods achieve this by explicitly disentangling attributes at the architectural or data level: separate embedding spaces for each attribute dimension, multi-head prediction, contrastive disentanglement losses, or CLIP-based prompt tuning.

Our approach differs fundamentally. We do not decompose classes into attribute pairs --- the model sees only flat string labels with no structural hint. If an external mechanism has already separated attributes, the resulting generalization reflects that externally imposed structure, not an internally induced one. We investigate whether structured label mixing alone, without architectural or data-level decomposition, can induce internal attribute separation.

\subsection{Label Smoothing}

Standard label smoothing \cite{label_smoothing_2024} uniformly redistributes probability mass from the true class to all other classes, serving as a regularizer that prevents overconfident predictions. Structure-aware variants such as SALS \cite{sals_2021} incorporate class relationships --- typically from external graphs or embeddings --- to produce non-uniform soft targets.

Our method also produces non-uniform soft targets, but derives them from string similarity between class names rather than from external semantic or structural knowledge. This is a deliberate modeling choice: the similarity signal mimics how natural language links categories through shared symbolic fragments, independently of the actual semantic relationships between the referents.

\subsection{Other Soft-Label Methods}

Mixup \cite{mixup_2018} and its extensions \cite{cumix_2020} create synthetic training examples by interpolating between images and their labels, augmenting the input space. Knowledge distillation \cite{hinton_distillation_2015} transfers soft target distributions from a trained teacher model to a student. Both produce soft labels, but from different sources: interpolation or a pre-trained model. Our soft targets are constructed directly from the syntactic structure of class names --- a fixed, deterministic transformation that requires no pre-trained model, no new images, and no learned semantic information.

\subsection{Noisy Labels and Latent Generalization}

Research on noisy labels predominantly focuses on robustness: detecting and mitigating the harmful effects of label corruption on generalization \cite{rolnick_2017_robust}. Deep networks can memorize datasets with substantial label corruption \cite{nishi_2021_augmentation}, and various strategies --- sample reweighting, loss correction, contrastive learning --- have been developed to train robustly despite noise.

A recent line of work reveals that even when networks memorize noisy labels, their intermediate representations retain latent generalization ability recoverable through probing \cite{latent_generalization_2026}. This finding resonates with our probe experiment (Section~VI), where a linear probe on frozen features recovers compositional generalization that the original classifier fails to express.

However, all these works treat label noise as a \textit{problem} to be mitigated. Our approach inverts this perspective: we use structured label distortion as a \textit{productive} mechanism that induces capabilities absent from undistorted training.

\subsection{Data Augmentation with False Samples}

Data augmentation typically assumes label preservation --- the augmented sample retains the original class \cite{nishi_2021_augmentation}. Our expanded dataset method deliberately violates this assumption: false copies of training images are added with incorrect but structurally motivated class assignments. This is closer to pseudo-labeling and self-training \cite{pseudo_labeling_review_2026}, but differs in that our false labels are not model-generated predictions --- they are deterministic transformations derived from an external similarity function, and they are \textit{intentionally} incorrect.

\section{Method}

We propose two complementary methods of structured data distortion, both derived from a single similarity function over class names. The first --- \textit{mixed labels} --- modifies the target vectors. The second --- \textit{expanded dataset} --- adds false training samples.

\subsection{Class Encoding}

We consider a classification task with $K$ classes, where each class represents a combination of two properties: color and shape. For $N_c$ colors and $N_s$ shapes, $K = N_c \times N_s$. Each class is assigned a string name by concatenating its color and shape with a delimiter: e.g., \texttt{red-circle}, \texttt{blue-triangle}.

Crucially, these names are treated as \textit{atomic labels} --- the model receives no indication that they encode a pair of independent properties. From the classifier's perspective, the $K$ classes form a flat, unstructured set. The compositional structure exists only in the label strings.

\subsection{Similarity Function}

We define a similarity function between two class names based on their shared string components, using a Jaccard-like measure over component character lengths. For class names $a$ and $b$, let $\text{parts}(a)$ denote the set of components obtained by splitting $a$ at the delimiter. The similarity is:

\begin{equation}
S(a, b) = \frac{\sum_{w \in \text{parts}(a) \cap \text{parts}(b)} |w|}{\sum_{w \in \text{parts}(a) \cup \text{parts}(b)} |w|}
\end{equation}

where $|w|$ is the character length of component $w$. For example, $S(\texttt{red-circle},\, \texttt{red-triangle}) = \frac{3}{3+6+8} = \frac{3}{17} \approx 0.176$ via shared \texttt{red}, while $S(\texttt{red-circle},\, \texttt{blue-triangle}) = 0$ since no component is shared.

The function is deliberately \textit{syntactic}, not semantic. It captures surface-level overlap in label strings, modeling the way natural language creates associations between categories through shared phonetic or symbolic fragments, independently of actual perceptual similarity.

\subsection{Mixed Labels}

In the standard setting, each training sample is assigned a one-hot target vector $\mathbf{y} \in \mathbb{R}^K$. For mixed training, we construct a soft target vector $\mathbf{\tilde{y}} \in \mathbb{R}^K$ for each sample with true class $j$:

\begin{equation}
\tilde{y}_i = \begin{cases}
S(c_j, c_i) & \text{if } i \neq j \\
1.0 & \text{if } i = j
\end{cases}
\end{equation}

where $c_i$ is the string name of class $i$. The true class retains value 1.0, while other classes receive non-zero values proportional to their name similarity with the true class. Mixed labels are normalized and used with a soft cross-entropy loss:

\begin{equation}
\mathcal{L}_{\text{soft}} = -\sum_{i=1}^{K} \hat{y}_i \log p_i, \quad \hat{y}_i = \frac{\tilde{y}_i}{\sum_{k=1}^{K} \tilde{y}_k}
\end{equation}

This loss encourages the model to distribute probability mass according to the mixed target distribution, encoding inter-class relationships derived from the similarity function. Note that $\mathcal{L}_{\text{soft}}$ with one-hot targets is equivalent to standard cross-entropy, so both standard and mixed training use the same loss function.

\subsection{Expanded Dataset}

The second distortion method operates at the data level. For each training image with true class $j$, we add false copies of the same image to the dataset with different class assignments. The probability of adding a false copy with class $i$ is:

\begin{equation}
P(\text{add false copy with class } i) = \beta \cdot S(c_j, c_i)
\end{equation}

where $\beta$ is the expansion rate controlling the proportion of false entries. For each candidate false class $i \neq j$ with $S(c_j, c_i) > 0$, a Bernoulli trial determines whether the copy is added. False copies receive standard one-hot labels for their assigned (false) class. Holdout classes are excluded from false assignments.

The expanded dataset thus contains the original images with correct one-hot labels, augmented by a fraction of the same images with incorrect but structurally motivated labels. This models, at the data level, the same linguistic associations that mixed labels encode at the target level: an image of a red circle appearing alongside the label ``red-triangle'' mirrors the way shared linguistic fragments create cross-category associations.

\section{Experimental Setup}

\subsection{Dataset}

\begin{figure}[t]
\centering
\includegraphics[width=0.85\columnwidth]{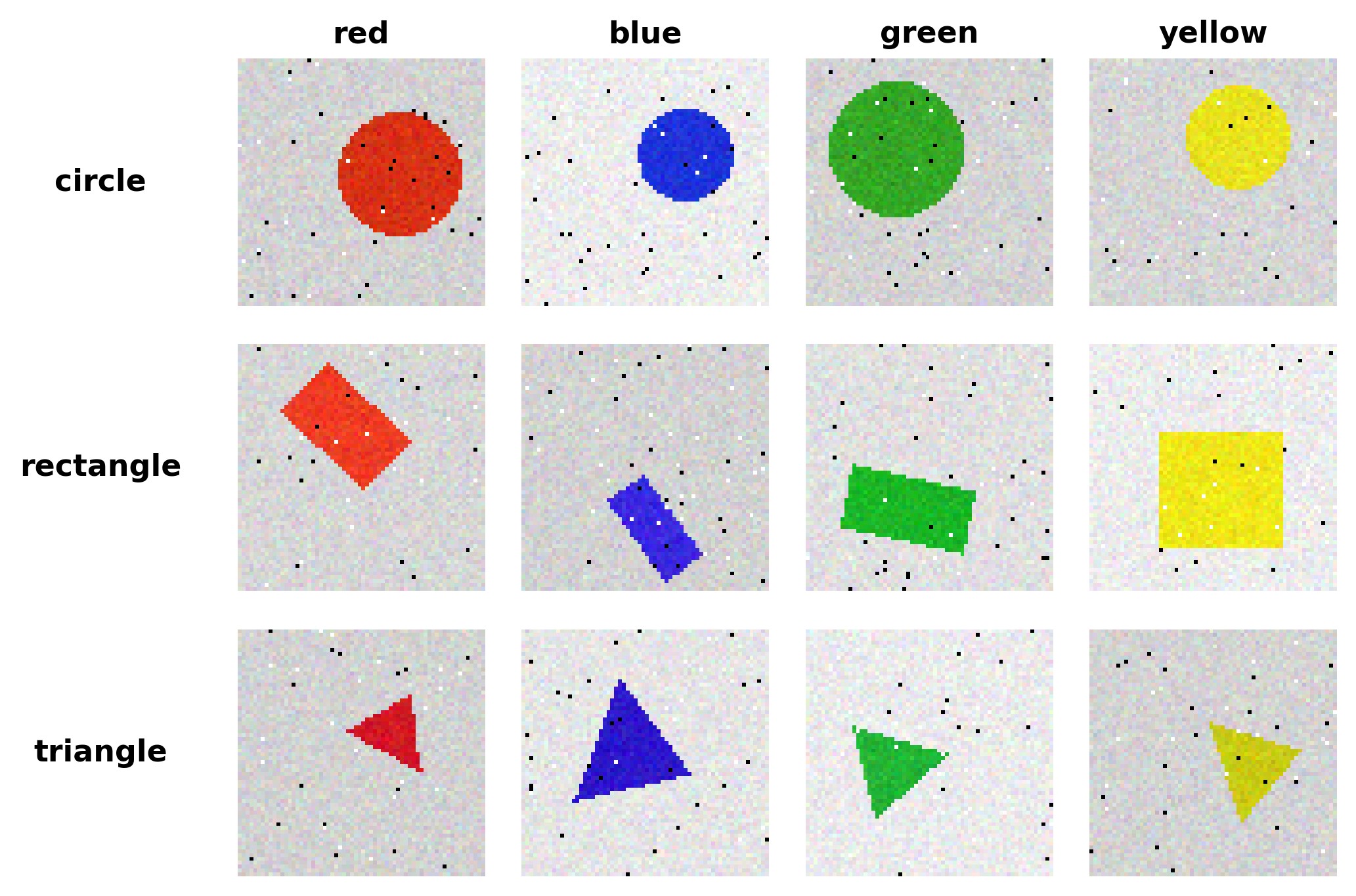}
\caption{Sample images for all 12 color-shape combinations (3 shapes $\times$ 4 colors).}
\label{fig:collage}
\end{figure}

We generate synthetic $64 \times 64$ RGB images, each containing a single geometric shape characterized by two properties: \textit{shape} (circle, rectangle, triangle) and \textit{color} (red, blue, green, yellow). Position, size, rotation, aspect ratio (for rectangles), and color shade are randomized. Gaussian noise and salt-and-pepper noise are added to all images. Fig.~\ref{fig:collage} shows representative samples.

The 3 shapes $\times$ 4 colors yield $K = 12$ class combinations. Randomized parameters and noise model natural variability beyond the two labeled properties.

\subsection{Data Splits}

Two class combinations are designated as \textit{holdout} and excluded from training entirely. The remaining 10 combinations are split into training (80\%) and validation (20\%) sets. The holdout set is used only for evaluation, testing whether the model can predict combinations it has never seen.

\begin{figure}[t]
\centering
\begin{subfigure}[t]{\columnwidth}
\centering
\includegraphics[width=0.85\columnwidth]{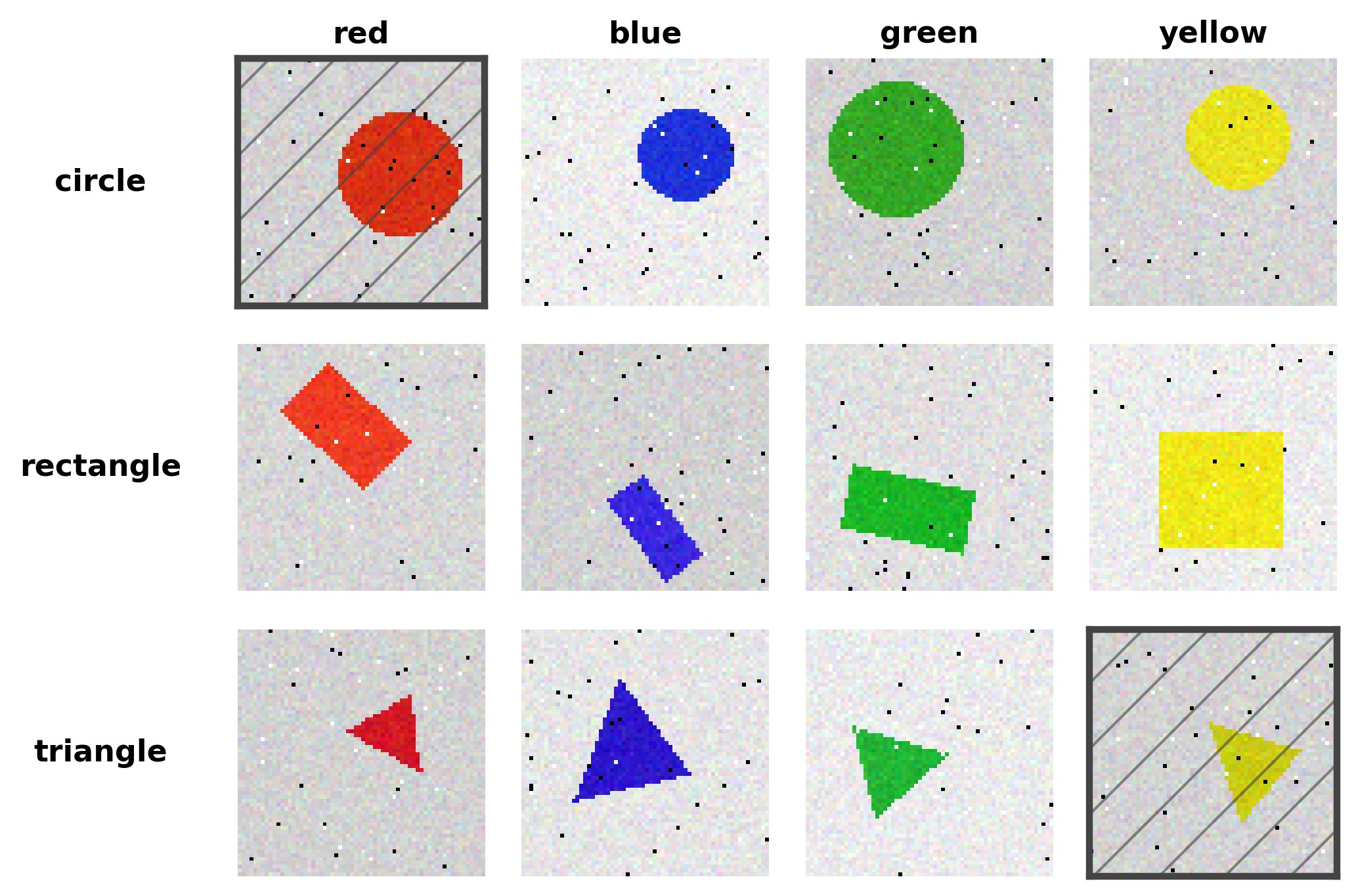}
\caption{Holdout A: red-circle and yellow-triangle}
\label{fig:holdout_a}
\end{subfigure}
\vspace{0.3cm}
\begin{subfigure}[t]{\columnwidth}
\centering
\includegraphics[width=0.85\columnwidth]{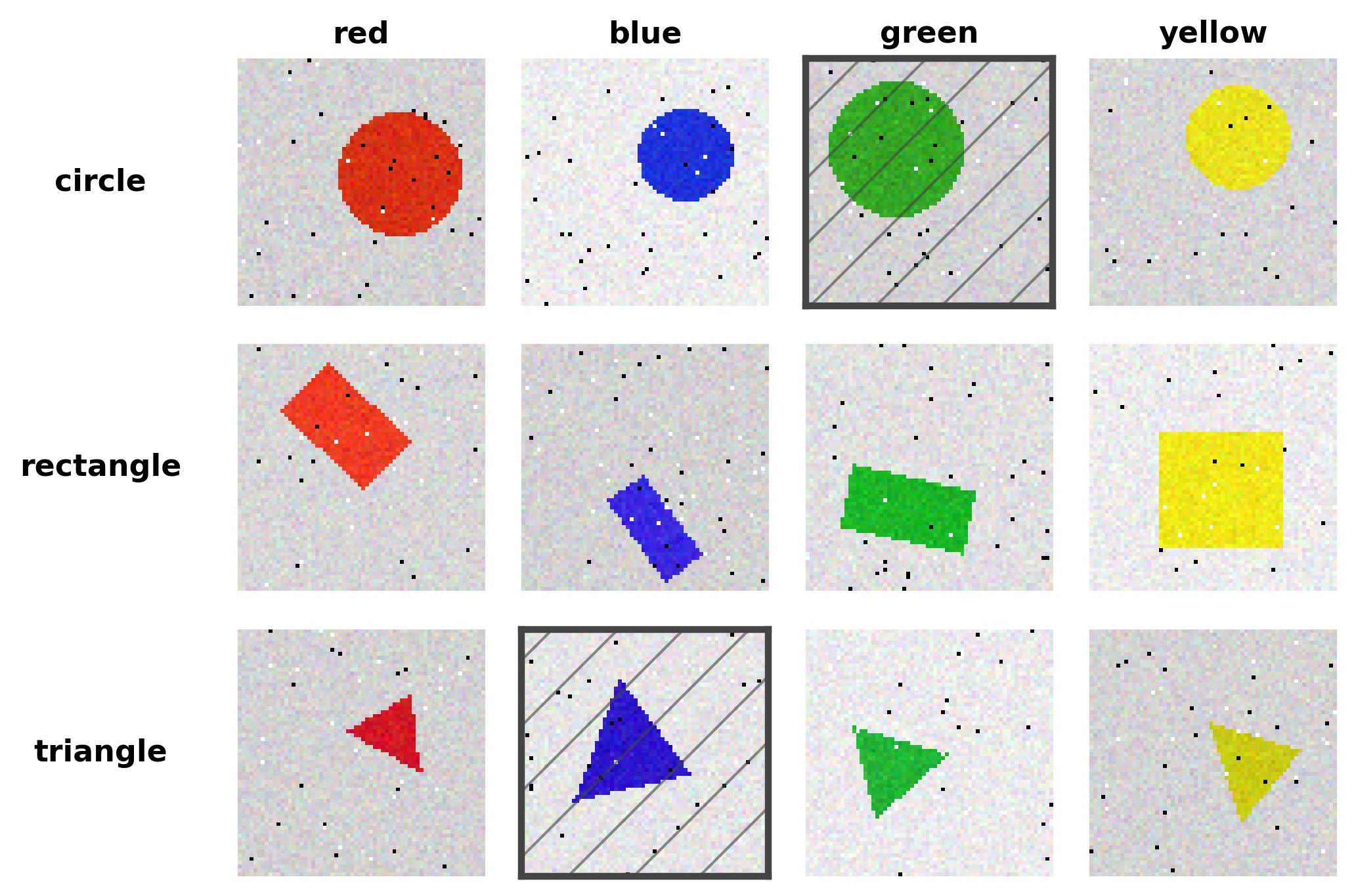}
\caption{Holdout B: green-circle and blue-triangle}
\label{fig:holdout_b}
\end{subfigure}
\caption{Holdout configurations. Hatched cells indicate class combinations excluded from training.}
\label{fig:holdouts}
\end{figure}

We evaluate two holdout configurations to test robustness: \textbf{Holdout~A} (red-circle + yellow-triangle) and \textbf{Holdout~B} (green-circle + blue-triangle), shown in Fig.~\ref{fig:holdouts}.

\subsection{Model Architecture}

\begin{figure}[t]
\centering
\includegraphics[width=\columnwidth]{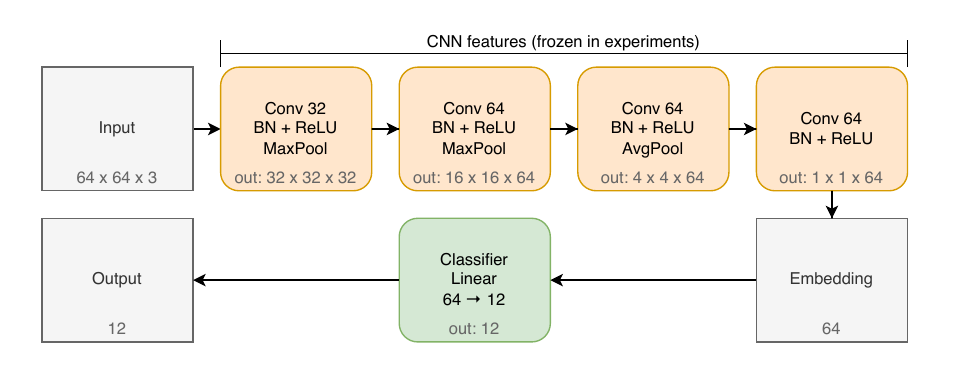}
\caption{Model architecture. CNN features (blue) produce a 64-dim embedding. The linear classifier (green) maps to $K$ classes. In probe experiments, CNN features are frozen and only the classifier is retrained.}
\label{fig:architecture}
\end{figure}

The architecture (Fig.~\ref{fig:architecture}) consists of two parts: a CNN feature extractor producing a 64-dimensional embedding, and a linear classifier.

The feature extractor contains four convolutional blocks. The first three blocks (32, 64, 64 channels) each apply convolution with $3 \times 3$ kernels, batch normalization, ReLU, and spatial reduction (max pooling or adaptive average pooling). The fourth block applies a $4 \times 4$ convolution that reduces the spatial dimensions to $1 \times 1$, producing a 64-dimensional embedding vector.

The classifier is a single linear layer ($64 \rightarrow K$) with no hidden layers or nonlinearities. This design choice is deliberate: a linear classifier can only separate classes by hyperplanes in the embedding space, which requires that the embedding be linearly separable with respect to all classes. This constraint makes the degree of factorization in the embedding space directly observable --- if holdout classes are correctly predicted, the embedding must place them in linearly separable regions, even though no holdout samples were seen during training.

\subsection{Training}

Models are trained for 50 epochs using Adam (learning rate $10^{-3}$) with ReduceLROnPlateau scheduling. Both standard and mixed training use the soft cross-entropy loss (Eq.~3), which is equivalent to standard cross-entropy when targets are one-hot. The model checkpoint with the lowest training loss is selected for evaluation.

Each configuration is repeated 50--100 times with fresh random initialization to assess reproducibility and produce distributional statistics.

\subsection{Evaluation}

All models are evaluated on the holdout set using per-class accuracy. Distributions of holdout accuracy across multiple runs are reported with 95\% binomial confidence intervals.

The complete experiment code is available online \cite{complexity_induction_code}.

\section{Mixed Training Results}

\subsection{Emergence of Compositional Generalization}

A model trained on standard one-hot labels achieves \textit{zero} holdout accuracy across all runs and both holdout configurations. The model learns to classify the 10 training combinations with near-perfect accuracy ($>$99\%) but cannot predict any unseen combination.

\begin{figure}[t]
\centering
\includegraphics[width=\columnwidth]{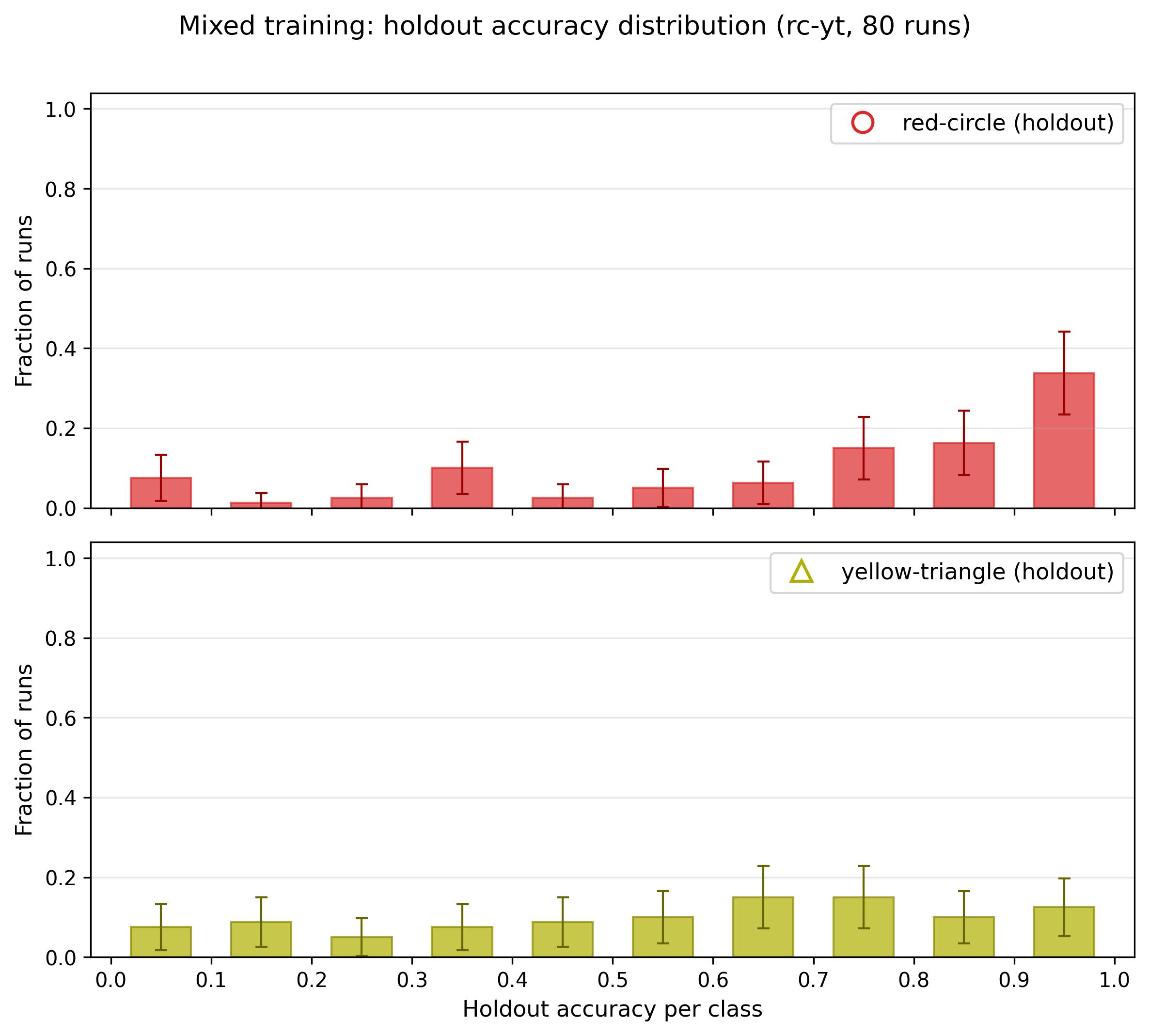}
\caption{Distribution of per-class holdout accuracy across 80 runs of mixed training (Holdout~A: red-circle + yellow-triangle). Error bars show 95\% CI.}
\label{fig:mixed_rcyt}
\end{figure}

When the same architecture is trained on mixed labels, holdout accuracy becomes non-zero. Fig.~\ref{fig:mixed_rcyt} shows the distribution of per-class holdout accuracy across 80 independent runs for Holdout~A (red-circle + yellow-triangle). Both holdout classes are predicted with non-trivial accuracy in the majority of runs, with average per-class accuracy of 70.8\% for red-circle and 56.7\% for yellow-triangle.

\subsection{Instability and Asymmetry}

Despite the consistent presence of the effect, two problems are apparent.

First, the holdout accuracy varies substantially across runs --- from below 0.1 to above 0.9 --- depending on random initialization. The qualitative property (non-zero holdout accuracy) is robust, but its magnitude is sensitive to initial conditions.

\begin{figure}[t]
\centering
\includegraphics[width=\columnwidth]{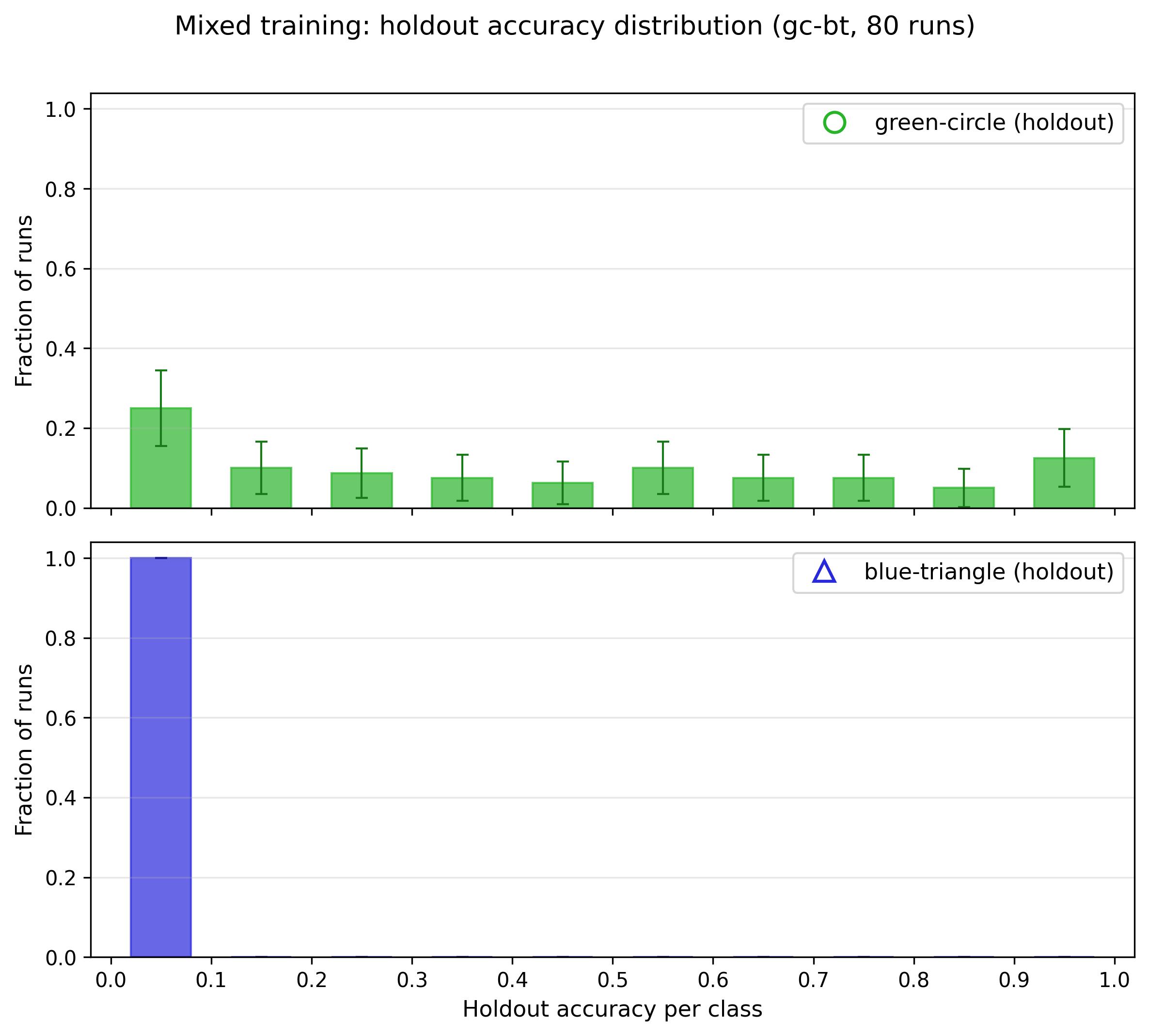}
\caption{Distribution of per-class holdout accuracy across 80 runs of mixed training (Holdout~B: green-circle + blue-triangle). Blue-triangle is almost never predicted.}
\label{fig:mixed_gcbt}
\end{figure}

Second, the effect is asymmetric across holdout configurations. Fig.~\ref{fig:mixed_gcbt} shows results for Holdout~B (green-circle + blue-triangle). While green-circle is predicted with moderate accuracy (avg.\ 41.9\%), blue-triangle is almost never predicted correctly (avg.\ 0.2\%). The same architecture and method that produces balanced results on Holdout~A fails on one class in Holdout~B.

These observations raise a question: where does the compositional generalization effect originate --- in the internal embedding representation, or in the classifier that maps embeddings to classes?

\section{Locating the Effect: Probe Experiment}

The instability and asymmetry of mixed training results suggest two possible explanations:

\begin{itemize}
    \item \textbf{Deep effect}: mixed labels restructure the embedding space, inducing factorization (separation of color and shape dimensions). The degree of factorization determines holdout accuracy.
    \item \textbf{Surface effect}: the CNN already factorizes the embedding during standard training. Mixed labels merely train the classifier to exploit this pre-existing structure for unseen combinations.
\end{itemize}

The linear classifier architecture enables a direct test. Since a single linear layer can only separate classes by hyperplanes, holdout prediction requires that holdout samples occupy linearly separable regions in the 64-dimensional embedding. If standard training already produces such an embedding, then the effect of mixed labels is limited to the classifier weights.

\subsection{Probe Design}

We test this with a two-phase probe experiment:
\begin{enumerate}
    \item Train the full model (CNN + classifier) on standard one-hot labels for 50 epochs. The resulting model achieves $>$99\% validation accuracy and 0\% holdout accuracy.
    \item Freeze all CNN weights. Re-initialize the classifier (a fresh linear layer $64 \rightarrow K$). Train only the classifier on mixed labels for 50 epochs.
\end{enumerate}

If the frozen CNN embedding already contains factorized structure, the new classifier should be able to find holdout classes in the embedding space, producing non-zero holdout accuracy.

\subsection{Probe Results}

The probe achieves substantial holdout accuracy on both configurations: avg.\ 0.773 on Holdout~A (red-circle 70.7\%, yellow-triangle 83.8\%) and avg.\ 0.412 on Holdout~B (green-circle 74.3\%, blue-triangle 10.4\%). These results are comparable to full mixed training, confirming that standard training already produces a partially factorized embedding. Mixed labels are not required for embedding factorization --- they activate the classifier to regions of the embedding space that correspond to unseen class combinations.

The asymmetry also persists: blue-triangle remains poorly predicted regardless of whether mixed labels are applied to the full model or only to the classifier. This indicates that the asymmetry originates in the embedding structure, not in the training procedure.

\section{Improving Factorization: Expanded Dataset}

Since the probe experiment shows that holdout accuracy is limited by embedding factorization, we ask whether the expanded dataset method (Section~III-D) can improve this factorization.

\subsection{Expanded Probe Design}

We repeat the probe experiment with one modification: in Phase~1, the model is trained on the expanded dataset ($\beta = 0.2$) for 200 epochs (longer training is required due to the contradictory gradient signals from false labels). Phase~2 remains identical --- freeze CNN, retrain classifier on mixed labels for 50 epochs. This isolates the effect of expanded training on the embedding.

\subsection{Comparison: Standard vs.\ Expanded vs.\ Random}

\begin{figure}[t]
\centering
\includegraphics[width=\columnwidth]{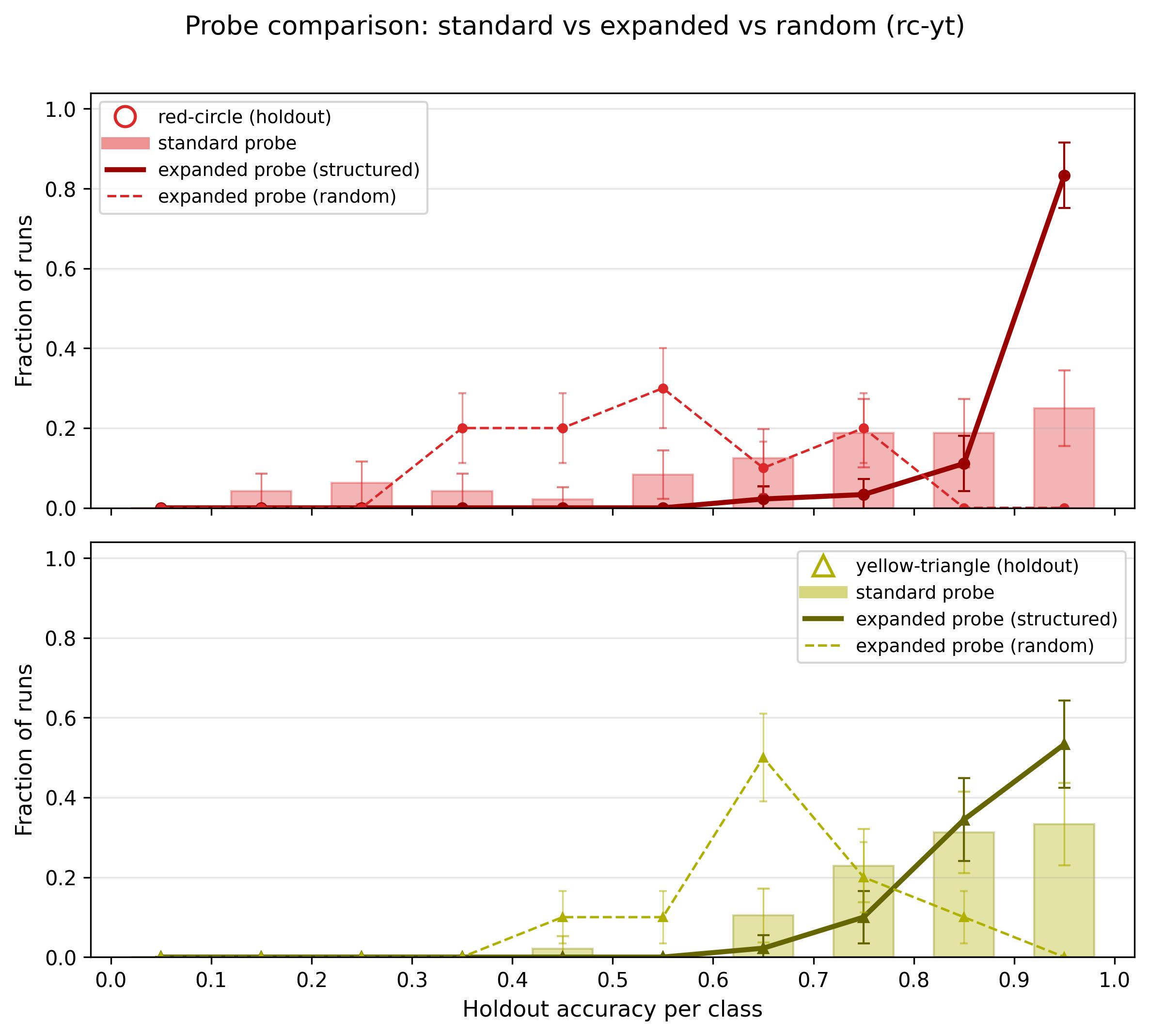}
\caption{Probe comparison for Holdout~A (red-circle + yellow-triangle). Bars: standard probe. Solid line: expanded (structured). Dashed line: expanded (random). Error bars: 95\% CI.}
\label{fig:probe_rcyt}
\end{figure}

\begin{figure}[t]
\centering
\includegraphics[width=\columnwidth]{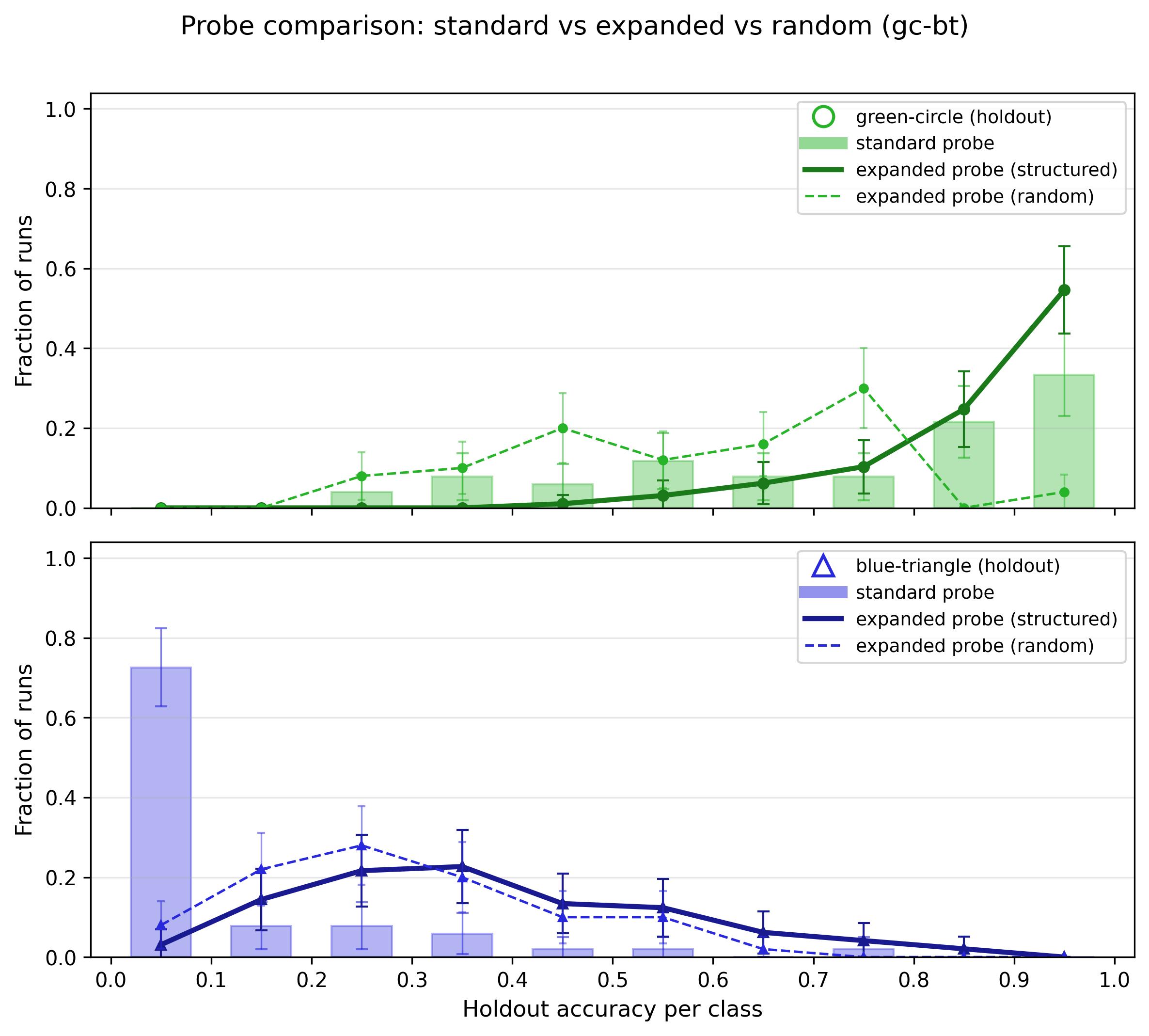}
\caption{Probe comparison for Holdout~B (green-circle + blue-triangle). Expanded (structured) shifts the distribution rightward for both classes, including the previously problematic blue-triangle.}
\label{fig:probe_gcbt}
\end{figure}

Fig.~\ref{fig:probe_rcyt} and Fig.~\ref{fig:probe_gcbt} compare three conditions across both holdout configurations:
\begin{itemize}
    \item \textbf{Standard probe}: Phase~1 on standard one-hot data
    \item \textbf{Expanded probe (structured)}: Phase~1 on expanded dataset with Jaccard-proportional false entries
    \item \textbf{Expanded probe (random)}: Phase~1 on expanded dataset with random class assignments (same number of false entries, but class chosen uniformly at random)
\end{itemize}

Table~\ref{tab:probe_comparison} summarizes the results.

\begin{table}[t]
\centering
\caption{Probe comparison across holdout configurations.}
\label{tab:probe_comparison}
\begin{tabular}{lccc}
\toprule
& \textbf{Standard} & \textbf{Expanded} & \textbf{Random} \\
& \textbf{probe} & \textbf{(structured)} & \textbf{(random)} \\
\midrule
\multicolumn{4}{l}{\textit{Holdout A: red-circle + yellow-triangle}} \\
\quad Avg.\ holdout acc. & 0.773 & \textbf{0.911} & 0.613 \\
\quad red-circle & 70.7\% & \textbf{93.5\%} & 55.4\% \\
\quad yellow-triangle & 83.8\% & \textbf{88.7\%} & 67.0\% \\
\midrule
\multicolumn{4}{l}{\textit{Holdout B: green-circle + blue-triangle}} \\
\quad Avg.\ holdout acc. & 0.412 & \textbf{0.614} & 0.434 \\
\quad green-circle & 74.3\% & \textbf{87.0\%} & 58.3\% \\
\quad blue-triangle & 10.4\% & \textbf{37.6\%} & 29.5\% \\
\bottomrule
\end{tabular}
\end{table}

The structured expanded dataset consistently improves holdout accuracy over the standard probe, with particularly notable gains on previously problematic classes: blue-triangle improves from 10.4\% to 37.6\%. The random expanded dataset, by contrast, performs \textit{worse} than the standard probe on most metrics --- random noise degrades the natural embedding factorization rather than enhancing it.

\subsection{Embedding Visualization}

\begin{figure}[t]
\centering
\begin{subfigure}[t]{\columnwidth}
\centering
\includegraphics[width=0.85\columnwidth]{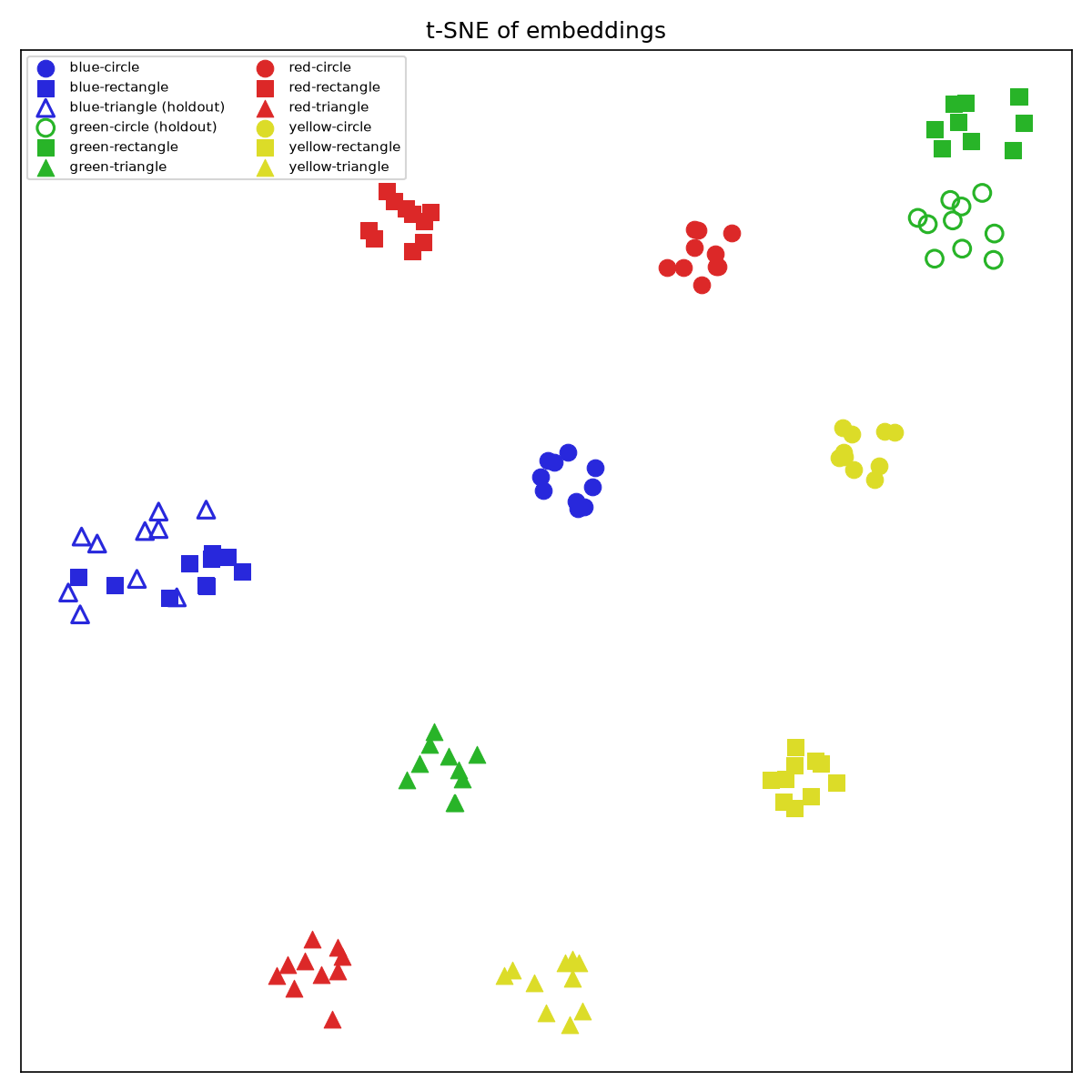}
\caption{Standard probe}
\label{fig:tsne_standard}
\end{subfigure}
\vspace{0.3cm}
\begin{subfigure}[t]{\columnwidth}
\centering
\includegraphics[width=0.85\columnwidth]{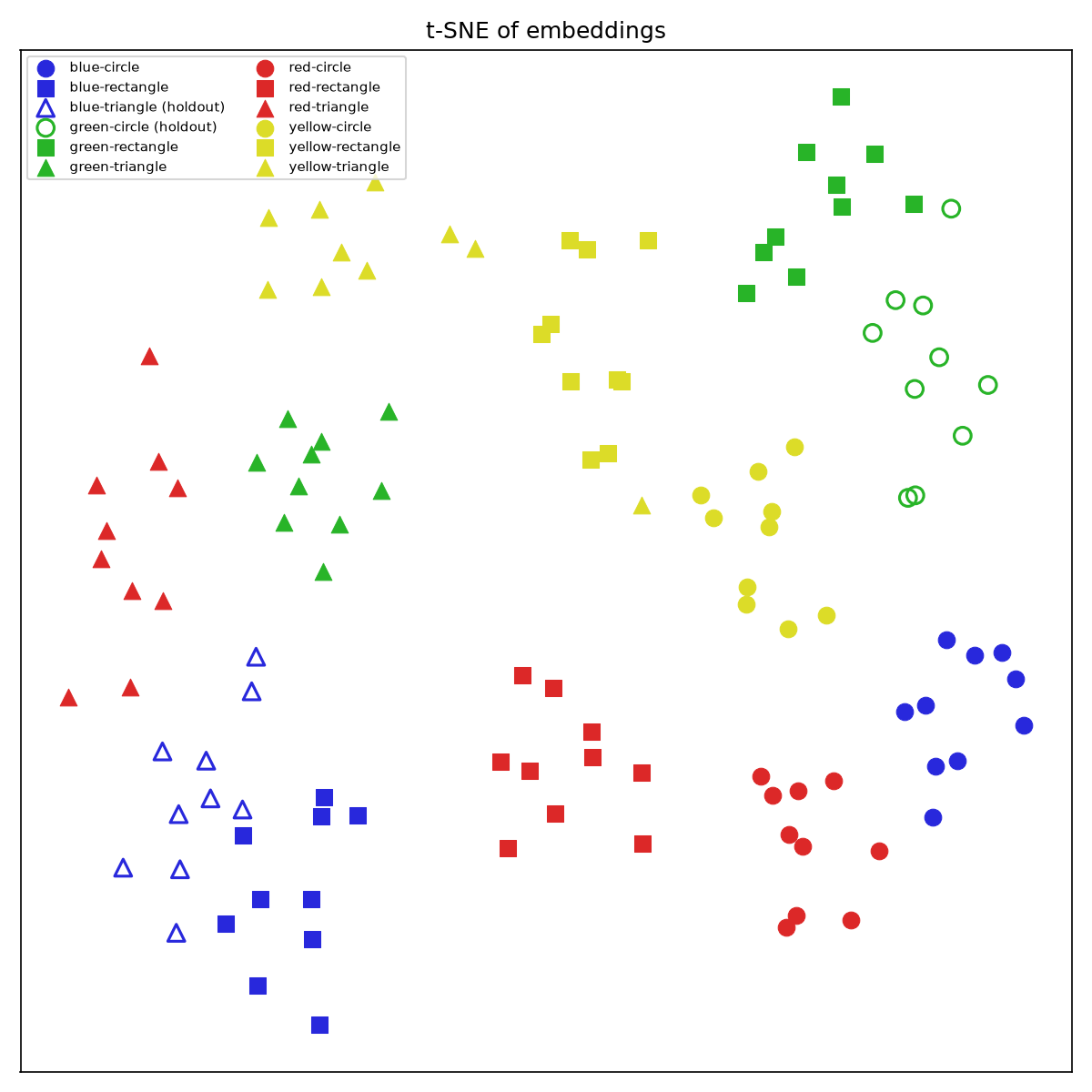}
\caption{Expanded probe (structured)}
\label{fig:tsne_expanded}
\end{subfigure}
\caption{t-SNE of embeddings (Holdout~B). Filled markers: validation classes. Hollow markers: holdout classes. The expanded model produces more systematic spatial separation.}
\label{fig:tsne_comparison}
\end{figure}

Fig.~\ref{fig:tsne_comparison} shows t-SNE projections of the 64-dimensional embeddings for standard and expanded probe models. The expanded model produces a more spatially distributed embedding, with holdout classes (hollow markers) positioned between their related training classes rather than collapsed onto a single neighboring cluster.

\section{Discussion}

\subsection{Two Mechanisms of Complexity Induction}

The experiments reveal that complexity induction operates through two distinct mechanisms, corresponding to the two distortion methods:

\textbf{Mixed labels} act on the classifier. The probe experiment (Section~VI) shows that a CNN trained on standard one-hot data already produces a partially factorized embedding --- the visual properties of the data naturally lead to internal representations where color and shape are somewhat separated. Mixed labels do not restructure this embedding; instead, they train the classifier to exploit the pre-existing factorization for unseen combinations, assigning non-zero weights to regions of the embedding space corresponding to holdout classes.

\textbf{Expanded dataset} acts on the embedding. Structurally motivated false labels create gradient pressure that prevents the CNN from collapsing visually similar classes into overlapping regions, producing a more distributed representation where holdout classes naturally fall into linearly separable areas (Table~\ref{tab:probe_comparison}). Crucially, the random expanded control confirms that this effect requires \textit{structured} distortion: random false labels degrade embedding factorization rather than enhancing it.

\subsection{Hypothesis: Language as Complexity Induction}

In human cognitive development, perceptual systems naturally separate visual features (color, shape, texture) --- analogous to the CNN's natural embedding factorization observed in our probe experiment. Our results suggest that language may act on both levels: strengthening the perceptual separation itself (as expanded training improves embedding factorization), and connecting this separation to a combinatorial category system (as mixed labels activate the classifier for unseen combinations). The child who hears ``red ball'' and ``red truck'' may both sharpen the internal distinction between color and shape, and acquire a combinatorial system where ``red'' links otherwise unrelated objects.

We hypothesize that this parallel is not merely an analogy. If structured distortion of training signals can induce compositional generalization in an artificial system, then natural language --- which systematically complicates the perceptual context through combinatorial, arbitrary, and partially misleading associations --- may play a structurally analogous role in biological cognitive development. Under this hypothesis, language is not merely a communication tool or a cognitive aid, but a \textit{structurally necessary} signal that transforms a perceptual modeling system into a compositional one.

The experiment presented here is a minimal demonstration: a simple CNN, synthetic data, a single similarity function operating at the level of surface-level string similarity. Natural language operates across multiple levels of structure (phonetic, morphological, syntactic, semantic), unfolds over developmental time with changing complexity, and interacts with architectures shaped by evolution to be receptive to such signals. This suggests several directions for amplifying the observed effect: identifying architectures that are more susceptible to complexity induction; developing training regimes that vary the type and intensity of distortion over time, analogous to developmental stages; and extending the principle to nested levels of structure, where the output of one level of complexity induction serves as input to the next.

\section{Limitations}

This work uses synthetic data with simple geometric shapes. Whether the effect transfers to natural images with richer visual structure remains an open question. The class space is small ($K = 12$); behavior at larger scales requires further exploration.

Only one CNN architecture was tested. The relationship between model capacity, architecture, and susceptibility to complexity induction has not been explored. The linear classifier constraint, while enabling the probe analysis, may limit the generality of the conclusions --- deeper classifiers might compensate for poor embedding factorization.

The holdout accuracy varies substantially across runs, and the per-class distribution is often asymmetric. While the qualitative effect is reproducible, the quantitative outcome depends on random initialization and the specific holdout combination.

The connection to natural language is presented as a motivating analogy. The Jaccard similarity function is a highly simplified model of linguistic associations and does not capture the full complexity of how language interacts with perception and learning.

\section{Conclusion}

We have demonstrated that structured distortion of training data can induce compositional generalization in a standard CNN classifier, and that this effect operates at two levels: structured false labels improve embedding factorization, while mixed soft targets activate the classifier for unseen combinations. Random distortion degrades rather than enhances performance, confirming that the structure of the distortion --- not noise per se --- is the operative factor. These findings point to a general principle: appropriately structured complication of training signals can produce qualitatively new generalization properties absent from undistorted training.

The complete experiment code is available online \cite{complexity_induction_code}.

\bibliographystyle{IEEEtran}
\bibliography{references}

\begin{thebibliography}{10}
\providecommand{\url}[1]{#1}
\csname url@samestyle\endcsname
\providecommand{\newblock}{\relax}
\providecommand{\bibinfo}[2]{#2}
\providecommand{\BIBentrySTDinterwordspacing}{\spaceskip=0pt\relax}
\providecommand{\BIBentryALTinterwordstretchfactor}{4}
\providecommand{\BIBentryALTinterwordspacing}{\spaceskip=\fontdimen2\font plus
\BIBentryALTinterwordstretchfactor\fontdimen3\font minus
  \fontdimen4\font\relax}
\providecommand{\BIBforeignlanguage}[2]{{%
\expandafter\ifx\csname l@#1\endcsname\relax
\typeout{** WARNING: IEEEtran.bst: No hyphenation pattern has been}%
\typeout{** loaded for the language `#1'. Using the pattern for}%
\typeout{** the default language instead.}%
\else
\language=\csname l@#1\endcsname
\fi
#2}}
\providecommand{\BIBdecl}{\relax}
\BIBdecl

\bibitem{czsl_survey_2025}
A.~Munir, F.~Z. Qureshi, M.~Ali, and M.~H. Khan, ``Compositional zero-shot
  learning: A survey,'' \emph{arXiv preprint arXiv:2510.11106}, 2025.

\bibitem{label_smoothing_2024}
H.~Ren, Y.~Zhao, Y.~Zhang, and W.~Sun, ``Learning label smoothing for text
  classification,'' \emph{PeerJ Computer Science}, vol.~10, p. e2005, 2024.

\bibitem{sals_2021}
Y.~Wang, Y.~Cai, Y.~Liang, W.~Wang, H.~Ding, M.~Chen, J.~Tang, and B.~Hooi,
  ``Structure-aware label smoothing for graph neural networks,'' \emph{arXiv
  preprint arXiv:2112.00499}, 2021.

\bibitem{mixup_2018}
H.~Zhang, M.~Cisse, Y.~N. Dauphin, and D.~Lopez-Paz, ``mixup: Beyond empirical
  risk minimization,'' in \emph{International Conference on Learning
  Representations (ICLR)}, 2018.

\bibitem{cumix_2020}
M.~Mancini, Z.~Akata, E.~Ricci, and B.~Caputo, ``Towards recognizing unseen
  categories in unseen domains,'' in \emph{European Conference on Computer
  Vision (ECCV)}, 2020.

\bibitem{hinton_distillation_2015}
G.~Hinton, O.~Vinyals, and J.~Dean, ``Distilling the knowledge in a neural
  network,'' \emph{arXiv preprint arXiv:1503.02531}, 2015.

\bibitem{rolnick_2017_robust}
D.~Rolnick, A.~Veit, S.~Belongie, and N.~Shavit, ``Deep learning is robust to
  massive label noise,'' \emph{arXiv preprint arXiv:1705.10694}, 2017.

\bibitem{nishi_2021_augmentation}
K.~Nishi, Y.~Ding, A.~Rich, and T.~H{\"o}llerer, ``Augmentation strategies for
  learning with noisy labels,'' in \emph{IEEE/CVF Conference on Computer Vision
  and Pattern Recognition (CVPR)}, 2021.

\bibitem{latent_generalization_2026}
S.~Ketha and V.~Ramaswamy, ``On the dynamics \& transferability of latent
  generalization during memorization,'' \emph{Transactions on Machine Learning
  Research (TMLR)}, 2026.

\bibitem{pseudo_labeling_review_2026}
P.~Kage, J.~C. Rothenberger, P.~Andreadis, and D.~I. Diochnos, ``A review of
  pseudo-labeling for computer vision,'' \emph{Journal of Artificial
  Intelligence Research (JAIR)}, vol.~85, 2026.

\bibitem{complexity_induction_code}
A.~Abramov, ``Complexity induction: Experiment code,''
  \url{https://github.com/avabr/complexity-induction}, 2026.

\end{thebibliography}

\end{document}